\documentclass{article}

\usepackage{arxiv}

\usepackage{amsmath,amsfonts,bm}

\def\eqref#1{equation~\ref{#1}}

\def\1{\bm{1}}

\DeclareMathAlphabet{\mathsfit}{\encodingdefault}{\sfdefault}{m}{sl}
\SetMathAlphabet{\mathsfit}{bold}{\encodingdefault}{\sfdefault}{bx}{n}

\DeclareMathOperator*{\argmin}{arg\,min}

\usepackage[utf8]{inputenc} 
\usepackage[T1]{fontenc}    
\usepackage{hyperref}       
\usepackage{url}            
\usepackage{booktabs}       
\usepackage{amsfonts}       
\usepackage{nicefrac}       
\usepackage{microtype}      
\usepackage{lipsum}
\usepackage[numbers]{natbib}
\usepackage{graphicx}
\usepackage{amsmath}
\usepackage{amsfonts}

\title{Embryo staging with weakly-supervised region selection 
and dynamically-decoded predictions}

\usepackage{authblk}

\makeatletter
\renewcommand\AB@affilsepx{ \protect\Affilfont}
\makeatother

\author[1]{Tingfung Lau}
\author[2]{Nathan Ng}
\author[3]{Julian Gingold}
\author[3]{Nina Desai}
\author[4]{Julian McAuley}
\author[1]{Zachary C. Lipton}

\affil[1]{Carnegie Mellon University} 
\affil[2]{Facebook}
\affil[3]{Cleveland Clinic Foundation}
\affil[4]{University of California San Diego}
\affil[ ]{\texttt{tingfunl@cs.cmu.edu, n.ng555@gmail.com, \{gingolj,desain\}@ccf.org, jmcauley@eng.ucsd.edu, zlipton@cmu.edu}}

\begin{document}

\maketitle

\begin{abstract}
To optimize clinical outcomes, 
fertility clinics must strategically select which embryos to transfer.
Common selection heuristics are formulas expressed in terms of the durations required to reach various developmental milestones,
quantities historically annotated manually
by experienced embryologists based on time-lapse \emph{EmbryoScope} videos. 
We propose a new method for automatic embryo staging
that exploits several sources of structure in this time-lapse data.
First, noting that in each image the 
embryo occupies a small subregion,
we jointly train a region proposal network with the downstream classifier to isolate the embryo.
Notably, because we lack ground-truth bounding boxes, 
our we weakly supervise the region proposal network 
optimizing its parameters via reinforcement learning 
to improve the downstream classifier's loss.
Moreover, noting that embryos reaching the blastocyst stage progress monotonically through earlier stages, 
we develop a dynamic-programming-based decoder that 
post-processes our predictions to select 
the most likely monotonic sequence of developmental stages.
Our methods outperform vanilla residual networks and rival the best numbers in contemporary papers,
as measured by both per-frame accuracy and transition prediction error, despite operating on smaller data than many.

\end{abstract}


\section{Introduction}
\label{sec:introduction}
Following its introduction in 1978, in vitro fertilization (IVF),
in which an egg and sperm are combined outside the body,
has rapidly emerged as one of the most successful assisted reproductive technologies,
 contributing to roughly 1.7\% of all births in the United States \citep{cdc2016}.
A single cycle of IVF may lead to the growth of multiple ovarian follicles, 
each of which may contain an oocyte (egg cell). 
These oocytes are aspirated with a fine needle using ultrasound guidance 
while the patient is under anesthesia and subsequently fertilized with sperm.
Only a fraction of the oocytes fertilize, 
and a smaller fraction continue to grow and develop normally as embryos 
before being considered ready for transfer into the uterus 
(typically after 5-6 days, though some labs use embryos grown only 3 days). 
Although this process typically generates multiple embryos, 
most infertility clinics strongly encourage (and some require) 
transfer of only one embryo at a time 
because of the greater maternal and fetal risks 
associated with multi-fetal gestation. 

Unfortunately, even under the best circumstances and with genetic testing, 
the implantation rate following embryo transfer is around 70\% \citep{simon2018pregnancy} 
and may be significantly lower without genetic testing, 
meaning that patients may be forced to undergo multiple transfers 
of embryos generated from IVF cycle(s) 
in order to achieve a single normal pregnancy. 
This leads to the clinical challenge of identifying and prioritizing 
those embryo(s) most likely to lead to a normal pregnancy with the fewest total transfers, 
in the least time and at the lowest cost. 
These priorities are often in direct conflict, 
leading to the wide variability in clinical decisions 
made between doctors, clinics and countries.



To prioritize embryo selection,
embryologists typically incorporate scores 
based on a morphological evaluation of each embryo \citep{cetinkaya2016morpho}. 
Historically, embryos were removed from the incubator 
for assessment under a microscrope 
by a trained embryologist one to two times daily.   
The development of incubators with built-in time-lapse monitoring 
has enabled non-invasive embryo assessment with comparatively fine-grained detail, 
inspiring significant interest in applying embryo ``morphokinetics'' 
to score and prioritize embryos \citep{chamayou2013morpho}. 
Informally, the morphokinetics comprise the timing 
and morphologic appearance of embryos as they grow 
and pass through a series of sequential developmental stages, 
with earlier stages corresponding to cell divisions 
and subsequent stages corresponding to larger structural milestones,
e.g.~formation of the blastocyst.

Modern incubators use a high-powered microscope 
to capture images of a developing embryo approximately every $15$ minutes.
Currently, embryologists must perform the morphokinetic analysis manually, 
viewing a sequence of photographs and annotating the time stamps 
at which each embryo achieves various developmental milestones.
These scores are combined according to heuristic formulas
to rank the embryos by their putative viability 
for transfer into a prepared endometrium. 

In this paper, we investigate machine learning techniques 
for automatically detecting these transition times.
Specifically, we propose several methods 
to (i) learn region proposal models from weak supervision 
to discard background so that the classifier 
can focus on the region corresponding to the embryo;
(ii) incorporate the temporal context of the video into the model architecture; 
and (iii) post-process our predictions at the sequence level, 
using dynamic programming to determine the most likely 
monotonically-increasing sequence of morphokinetic stages.

%
%

Our experiments focus on a dataset consisting of $1309$ 
time-lapse videos extracted from EmbryoScope\texttrademark (Vitrolife, Sweden) incubators 
at a large academic medical center's fertility clinic.
Each frame in the raw videos has $500\times500$ resolution.
We downsize these to the standard $224\times224$ ImageNet dimensions
for compatibility with pretrained nets and corresponding hyperparameters. 

Compared to a baseline deep residual network (ResNet) \citep{he2016deep},
we find significant benefits from each of our three proposed techniques. 
Our region proposal network selects a $112\times112$ region 
and is optimized by reinforcement learning, 
following the policy gradient algorithm, 
using the cross entropy loss of the downstream classifier as a reward signal.
This technique improves frame-level accuracy from $83.65$ to $86.34$.
Adding a Long Short-Term Memory (LSTM) recurrent neural network
to post-process the predictions, we achieve an additional gain, 
boosting frame-level accuracy to $88.28\%$. 
Finally, we evaluate two variants of our dynamic programming technique 
for decoding monotonic predictions, 
one based on finding the \emph{most likely} monotonically increasing  sequence, 
and another that minimizes the expected distance 
between the predicted and actual states 
(e.g.~when the ground truth is 
`stage $4$', we prefer to predict stage $3$ over stage $2$).
When applied to the raw classifier, the dynamic programming post-processing
methods confer improvements to frame-level accuracy of
$85.54\%$ and $86.34\%$, respectively.
Notably, our techniques yield complementary benefits:
altogether, they combine to achieve a frame-level accuracy of
$90.23\%$,
and reduce the transition-level error (number of frames off) of the raw classifier by 
$30\%$ to 
$4.37$.

\section{Dataset}
\label{sec:dataset}

\begin{figure*}[t]    
  		\centering	\includegraphics[width=\linewidth]{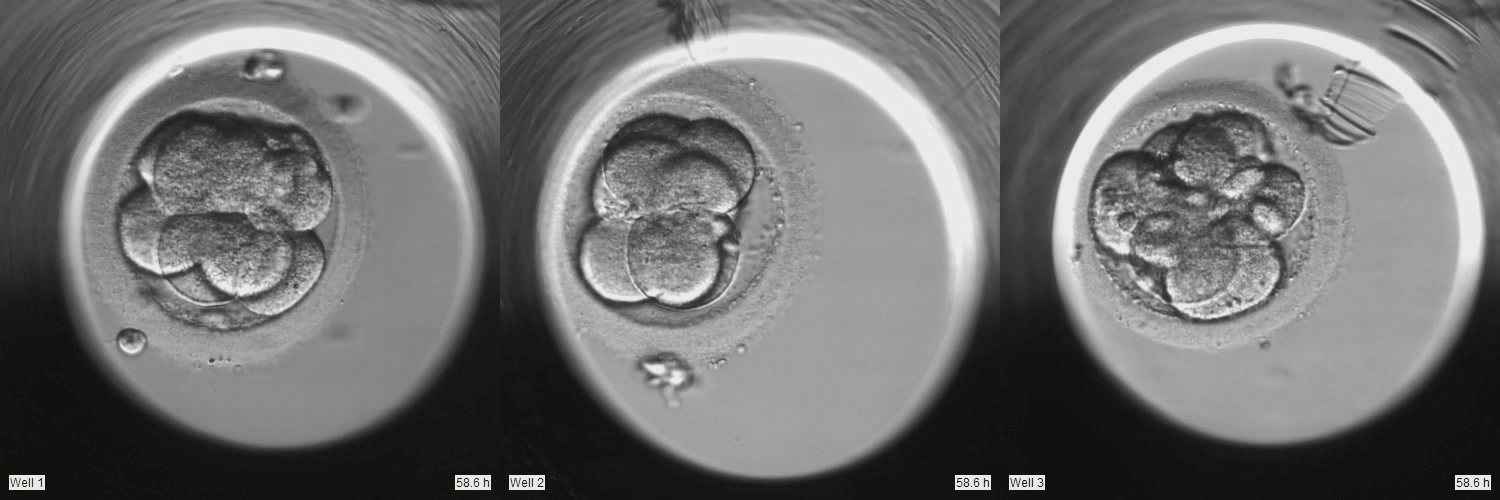}
		\caption{A sample frame containing 3 wells in an EmbryoScope video from our dataset.}
	\label{fig:video-ex}
\end{figure*}

The EmbryoScope time-lapse system is an embryo incubator 
capable of holding up to 12 wells simultaneously, each containing one embryo. 
Built in to the device are both a high-powered microscope 
and a camera used jointly to photograph each embryo on a $15$-minute cycle.
Each frame of the resulting time-lapse video 
consists of a $500\times500$ resolution grayscale image 
with a well number in the lower-left corner 
and the time superimposed in the lower-right corner, 
as seen in Figure \ref{fig:video-ex}.
The system also captures each image in multiple focal planes, 
although the central focal plane alone is used in this study.

\begin{figure*}[htbp]    
  		\centering	\includegraphics[width=1.0\linewidth]{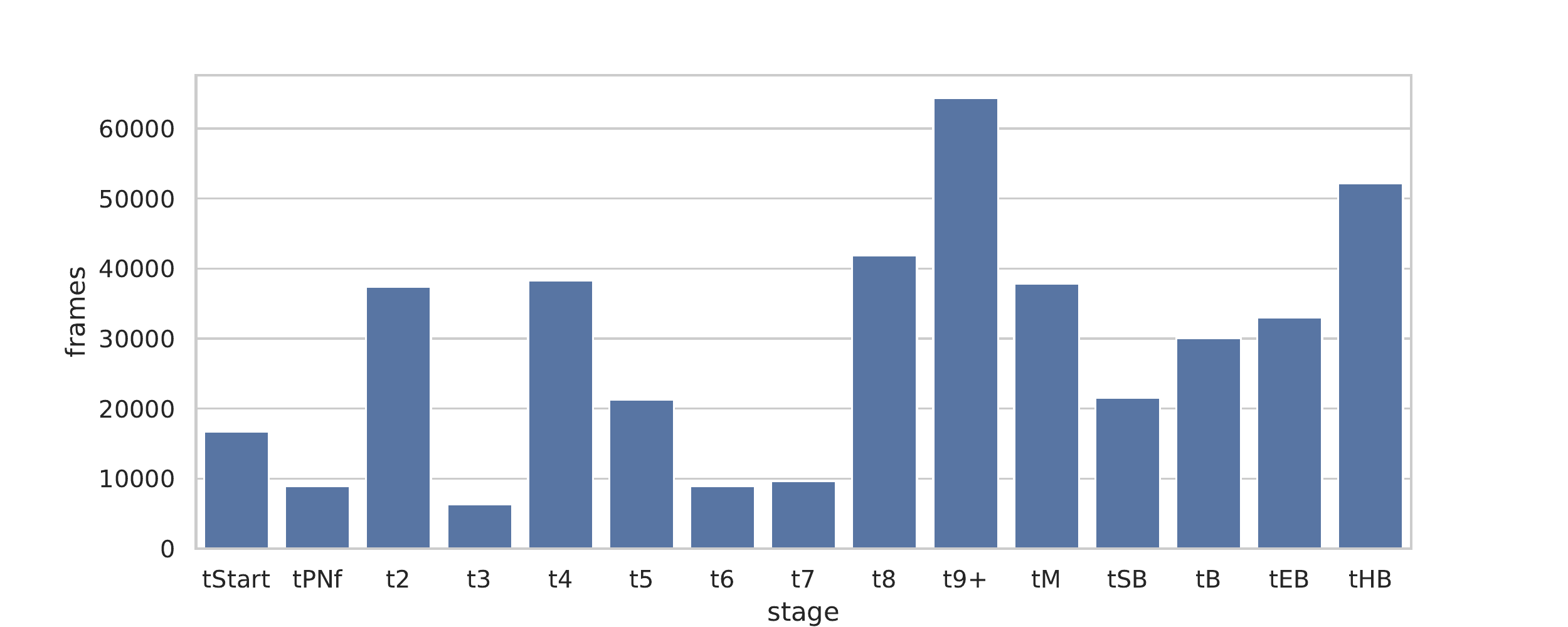}
		\caption{Distributions of morphokinetic stages across all labeled video frames.}
	\label{fig:stage-histo}
\end{figure*}

Our dataset consists of $1309$ EmbryoScope time lapse videos 
extracted from incubators
at a large academic medical center.
These videos span $113$ different patients, each with $11$ to $12$ wells 
and corresponding videos. 
Videos begin roughly $18$ hours after fertilization, 
and end roughly $140$ hours after fertilization. 
Annotations in the videos correspond to $15$ distinct morpho-kinetic stages, 
with the embryologist marking the time at which each embryo 
was first observed in each developmental stage.
Among cells that mature successfully (the rest are discarded),
the stages are monotonically increasing, 
meaning that among non-discarded embryos the ground truth labels 
never regress from a more advanced stage to a less mature stage.
We transform our stage transition annotations 
into per-frame stage labels by applying the most recently assigned stage.
We also assign a special \textbf{tStart} stage label 
to frames before the first stage.
The first observed stage corresponds to the moment 
when two pronuclei are visible (\textbf{tPNF}), 
and the next several stages correspond to cell divisions.
After the embryo reaches $8+$ cells, 
the subsequent stages correspond to higher-level features,
like the formation of the blastocyst. 
The embryo stage distribution in the full data set is given in Figure \ref{fig:stage-histo}.

Because most embryo selection heuristics depend 
only on the time to reach the cell division milestones,
in this study, we focus on the first six stages of development 
for each embryo, cutting off each video at hour $60$.
Moreover, these stages admit a cleaner problem  
because for a significant portion of our videos, 
expert (ground truth) annotations are missing for the latter $9$ stages. 
The $6$ stages that we address include the initial stage (\texttt{tStart}), 
the appearance and breakdown of the male and female pronucleus (\texttt{tPNf}), 
and the appearance of 2 through 4+ cells (\texttt{t2, t3, t4, t4+}).   
Among these frames, the class distribution 
is 11.48\%, 6.11\%, 25.70\%, 4.36\%, 25.85\%, 26.50\%.  
Because we want to be sure that our models generalize not only across frames,
or even across embryos but also across patients (mothers),
we stratify the dataset by patient, creating training/validation/test splits 
by randomly selecting $93$/$10$/$10$ patients and their respective wells. 
This yields $510$/$60$/$56$ embryos in the respective splits,
corresponding to $117553$/$14606$/$13305$ frames.

\section{Methods}
\label{sec:methods}



We cast predicting embryo morphokinetics as a multiclass classification problem, 
where the input is a time-lapse EmbryoScope\texttrademark video, 
and the output is a sequence of labels indicating 
the predicted stage of the embryo at each frame of the video. 

Our simplest method consists of applying standard image recognition
tools to predict the stage label $y_t$ for each frame given the image $x_t$. 
Image classification is now a mature technology,
and for all known related tasks, the current best-performing methods
are deep convolutional neural networks (CNNs).
All of our approaches are based upon convolutional neural networks.
Specifically, we choose the ResNet-50 architecture as our base model
due
to \citep{he2016deep}.
By default, this model takes as input a $224 \times 224$ resolution image 
which we downsize from the original $500 \times 500$ image.
The output consists of a $6$-dimensional softmax layer
corresponding to the $6$ class labels,
and we optimize the network in the standard fashion to minimize cross entropy loss.
We initialize the network with pre-trained weights learned 
on the benchmark ImageNet image recognition challenge \citep{russakovsky2015imagenet},
a practice widely known to confer significant transfer learning benefits
\citep{yosinski2014transferable}.
We suspect that given a more relevant source task with a comparably-large dataset
(ideally, concerning gray-scale images from cellular microscopy),
we might get even greater benefits, although we leave this investigation for future work.

\subsection{Weakly-Supervised Embryo Detection}

Motivating our first contribution for improving performance of the embryo classifier,
we observe that the embryo's cell(s) lie in a small region of the image,
and that the rest of the image, containing the rest of the well
and surrounding background consists only of imaging artifacts 
that have no relevance to stage prediction. 
We postulate that by first detecting where the embryo is, 
and then subsequently basing classifications 
on the cropped region containing only the cell,
we could filter out the background noise, improving predictive performance.
Moreover, since the subsequent classification is based on a smaller region, 
we could either (i) save computation, or 
(ii) refer back to the original image to extract a higher-resolution zoom 
on the cropped region, providing greater detail to the classifier.

The most standard way to cast the bounding box detection task
is to train a model with labeled data corresponding to the height and width of the box
as well as an $x$ and $y$ coordinates to locate the box.
For typical detection tasks, current deep learning-based object detection systems 
require large annotated datasets with bounding box labels.
However, we do not have any such labels available for our task.

To learn embryo-encapsulating bounding boxes without explicitly annotated boxes,
we propose a new approach that relies only on image-level class labels,
optimizing the region proposal model via weak supervision using reinforcement learning.
To begin, noting that the embryo size does not vary much,
we fix the box dimensions to $112\times112$ (a $.5\times.5$ crop),
focusing only on identifying the box center.
Since we only have the image-level label for the image classification task, 
the training objective of the detector $G$ 
is to help a downstream classifier $F$ to better classify the image. 
Our two-step detect-then-classify algorithm is described below:
\begin{enumerate}
    \item Given an input image $x$, the detector predicts a probabilistic distribution $G_{\theta}(\;\cdot\;;x)$ over a $14\times14$ rectangular grid of candidate box centers.
    \item Sample a region $R \sim G_{\theta}(x)$ and get the cropped subregion $x_R$.
    \item Feed $x_R$ to the classifier to predict probabilities for each class $F_{\phi}(x_R)$.
\end{enumerate}
Let $y$ be the label and $\ell$ be the usual cross entropy loss function. 
The expected loss of the two-step classification algorithm is 
\begin{equation}
    \mathcal{L}_{\mathit{cls}}(\theta,\phi) = \mathbb{E}_{R \sim G_{\theta}(\;\cdot\;;x)} [\ell(F_{\phi}(\mathrm{crop}(x, R)), y)]
\end{equation}
Note that both the detector $G$ and classifier $F$ share the objective of minimizing
the expected classification loss
$\mathcal{L}_{\mathit{cls}}$. 
The intuition behind this objective is that if the image crop 
has a larger intersection with the cell, 
it is easier for the classifier to classify the image. 
On the other hand, if a large part of the cropped image is background, 
the classifier should not perform much better than random guessing.
Note that our detector outputs a probability distribution over grid-cells.
At test time we make predictions by centering the bounding box 
at the expected $x$ and $y$ coordinates.


\begin{figure}[htbp]
    \centering
    \includegraphics[width=\textwidth]{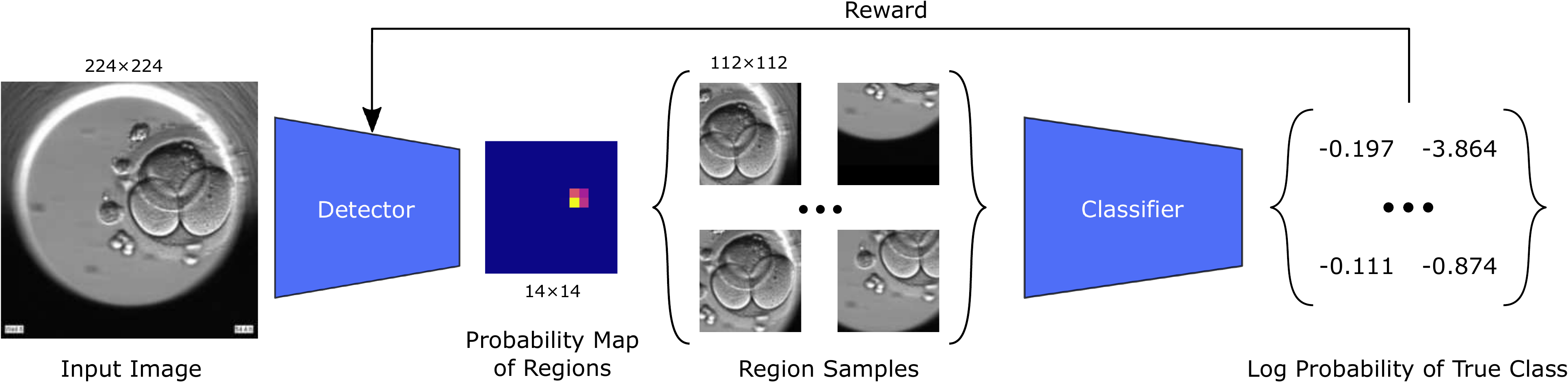}
    \caption{The pipeline of learning a detector in a weakly-supervised way using reinforcement learning.}
    \label{fig:detector-learn}
\end{figure}

The loss function involves computing the expectation with respect to all possible regions.
We use the Monte Carlo method to estimate the loss by drawing $K$ sample regions $R_1,R_2,\cdots,R_K \sim G_{\theta}(\;\cdot\;;x)$. 
The optimization problem becomes
\begin{equation}
    \argmin_{\theta,\phi} 
    \sum_{i=1}^{K} 
    \ell(F_{\phi}(\mathrm{crop}(x, R_i)), y).
    \label{eqn:obj}
\end{equation}
The gradient for the classifier's parameter $\phi$ is 
\begin{equation}
    \nabla_\phi \mathcal{L}_{\mathit{cls}}(\theta, \phi) = \frac{1}{k} \sum_{i=1}^{k} \nabla_{\phi} \ell(F_{\phi}(\mathrm{crop}(x, R_i)), y).
\end{equation}
The gradient for the detector's parameter $\theta$ 
is estimated using the policy gradient, a common reinforcement learning algorithm. 
Moreover, we incorporate a standard technique for variance reduction, 
use average rewards as a baseline $b$. This gives us the gradient for $\theta$, 
\begin{align}
    b &= \frac{1}{k}\sum_{i=1}^{k}  \ell(F_{\phi}(\mathrm{crop}(x, R_i)), y) \\
    \nabla_\theta \mathcal{L}_{\mathit{cls}}(\theta, \phi) &= \frac{1}{k} \sum_{i=1}^{k}  \left[\ell(F_{\phi}(\mathrm{crop}(x, R_i)), y) - b\right] \nabla_\theta \log G_\theta(R_i; x).
\end{align}
In preliminary experiments, we found that 
solely relying on the objective (\ref{eqn:obj}) 
converges quickly to an unsatisfactory local optimum
where the distributions of regions $G_\theta(\;\cdot\;;x)$ 
are always peaked on one specific region proposal. 
To overcome this issue, we encourage exploration 
in the reinforcement learning objective
by adding the negative entropy of the region distribution $G_\theta(\;\cdot\;;x)$,
a technique made by \citet{mnih2016asynchronous}.
The augmented overall loss function is 
\begin{equation}
    \mathcal{L}(\theta, \phi) = \mathcal{L}_{cls}(\theta, \phi) - \lambda H(G_\theta(\; \cdot \;; x))
\end{equation}
where $\lambda$ is the weight to balance the term.

\paragraph{Network structure} 
The detector predicts the region distribution 
using a sliding window method based on a Region Proposal Network (RPN) \citep{ren2015faster}. The region proposal is computed from a $14 \times 14$ intermediate feature map at \texttt{conv4\_2} in Resnet-50.
Based on our exploration of the data, 
we found that the embryo is typicaly contained 
in a rectangular region that is roughly one quarter the size of the image.
To simplify the distribution, we fix the width and height of the rectangle region 
to be $50\%$ of the size of the image using this prior knowledge. 
We assume that the center of the region proposal lies on
$14 \times 14$ grid, 
so that we only need to predict the probability of the region 
lying at each position in that grid. 
The probability is computed by applying a $3\times 3$ convolutional filter $W$ to the feature map followed by a softmax operation
\begin{equation}
    \gamma = \mathrm{softmax}(W * X_{\texttt{conv4\_2}}),
\end{equation}
where $\gamma_{ij}$ indicates the probability of selecting the box center at the $i$-th row and $j$-th column of the $14 \times 14$ grids.

Our base classifier is a Resnet-50 convolutional network
that takes a $112 \times 112$ cropped image as its input. 
We remove the layers in $\texttt{conv5}$ to speed up computation.

\paragraph{Test time} 
While the detector outputs a distribution of regions, 
at test time we want to use only ``the best'' region.
Some early experiments revealed the heuristic of choosing
the expected center coordinates of the predicted distribution. 
The average box center $(\bar{c}_x , \bar{c}_y)$ is computed by
\begin{align}
    \bar{c}_x &= \mathbb{E}_{c_x \sim G_\theta(c_x, c_y;x)} [c_x] \\
    \bar{c}_y &= \mathbb{E}_{c_y \sim G_\theta(c_x, c_y;x)} [c_y].
\end{align}

\subsection{LSTM}
Our first idea to incorporate context across adjacent frames 
is to employ recurrent neural networks with Long Short-Term Memory (LSTM) \citep{hochreiter1997long} units.
The LSTM takes as input a sequence of inputs, 
updates its internal state at each time, 
and predicts a sequence of outputs. 
The inputs to the LSTM consist of  $2048$-dimensional feature vectors 
extracted from the hidden layers of a vanilla CNN.
We then feed the feature vector to a bi-directional LSTM layer 
with $100$ units for each direction. 
We apply a linear mapping of the LSTM output at each time step to $6$ classes 
to get a sequence of predictions $y^{(1)}, y^{(2)}, \cdots, y^{(T)}$.
We set $T$ to $9$ optimizing the model to predict accurate on the middle $5$ frames.
We do not use predictions made on the first $2$ or last $2$ frames because they lack sufficient context.
\subsection{Structured Decoding with Dynamic Programming (DP)}
For embryos that successfully reach the blastocyst stage,
ground truth stages in our selected data set are monotonically non-decreasing, 
reflecting the condition that any viable embryo 
must continue to grow and developrather than arrest and die. 
The predictions of frame level CNNs or LSTMs with short sequences 
cannot learn this constraint since the model does not have enough context.
Therefore we impose this inductive bias
through a dynamic programming decoder 
that enforces monotonicity of predictions.
For each video, our model predicts the probability of the embryo stages 
$\hat{p}^{(t)} \in \mathbb{R}^6$
at every frame $t = 1,\cdots,T$,
where $T$ is the total number of frames in the video.
We want to find a decoded label sequence $\hat{\mathbf{y}}$ 
such that $\hat{y}^{(t+1)} \ge \hat{y}^{(t)}$ and $y^{(t)}$ 
most match the frame prediction $\hat{p}^{(t)}$ for each frame.
We define a potential function $\phi(y^{(t)},\hat{p}^{(t)})$ 
to measure how much the decoded label $\hat{y}^{(t)}$ 
deviates from $\hat{p}^{(t)}$ and turn the decoding 
to the following optimization problem:
\begin{equation}
    \argmin_{\hat{\mathbf{y}}} \sum_{t=1}^T \phi(\hat{y}^{(t)}, \hat{p}^{(t)}) \quad \mathrm{s.t.} \quad \hat{y}^{(t+1)} \ge \hat{y}^{(t)} \; \forall \; t \in \{1, \cdots, T-1\}.
\end{equation}
We investigate two potential functions, 
the negative log likelihood (NLL) and the earth mover's distance (EMD),
defined by $\phi_{\textit{NLL}}(\hat{y}, \hat{p}) = -\log(\hat{p}_{\hat{y}})$ 
and $\phi_{\textit{EMD}}(\hat{y},\hat{p}) = \sum_{s=1}^{S} \hat{p}_s |\hat{y} - s|$,
respectively, where $S=6$ is the number of development stages.
This optimization problem can be solved in polynomial time
using Dynamic Programming (DP) with a forward pass and a backward pass.

\section{Experiments}
\label{sec:experiments}

\subsection{Embryo Detection}
We train the region proposal network 
with SGD with momentum $0.9$ and learning rate $0.01$, with batch size set to $16$. 
The image is first downsampled to $224 \times 224$ before feeding into the detector. 
For each image in the batch, we sample $10$ regions, 
extracted as $112 \times 112$ images cropped the from $224 \times 224$ input image, 
and feed the cropped images into the classifier. 
We train two detectors with and without entropy regularization 
($\lambda = 0.01, 0$ respectively) 
to measure the effect of using the augmented loss function.

We also compare another approach to learn the detector 
using differentiable bi-linear sampling. 
The idea is that the detector only predicts
a single region that is fed to the classifier. 
We use differentiable bi-linear sampling 
when cropping the image at that region 
so that the gradient with respect to the classification loss 
can be back-propagated to the detector.
We change the last layer of the detector 
to be a fully connected layer to predict 
the coordinates $c_x,c_y$ of the center of the box. 
We were unable to make this alternative approach converge using SGD, 
so we eventually settled on the Adam optimizer with default parameters.

To evaluate the performance of the learned detector, 
we manually label a tiny data set with 120 images 
randomly sampled from the validation set, 
corresponding to 20 images from each embryo stage, 
and use these ground truth labels to get a quantitative evaluation of the detector. 
We report the Jaccard index, which is calculated 
by the intersection over union between the ground truth box and predicted box, 
as well as the euclidean distance between the ground truth box center 
and predicted box center, measured in pixels in the $500 \times 500$ raw image.
We also include the classification accuracy 
of a two-step detector-classifier on the selected $120$ images 
as this is our actual training objective. 

Detection results are shown in Table~\ref{tab:detection}. 
Our RL training with
entropy loss achieves a Jaccard index of $0.6957$
and a center distance only $11.58$ pixels from the manually labeled $500 \times 500$ images. 
The detector trained without the entropy term underperforms the detector with entropy, 
reflecting network convergence to some local optimum based on the current best performing region at an early stage. 
The differentiable sampling approach performs poorly for detection;
this shows that using a stochastic region proposal 
in our RL training is crucial for successfully training a detector.

\begin{figure}[htbp]
    \centering
    \includegraphics[width=\textwidth]{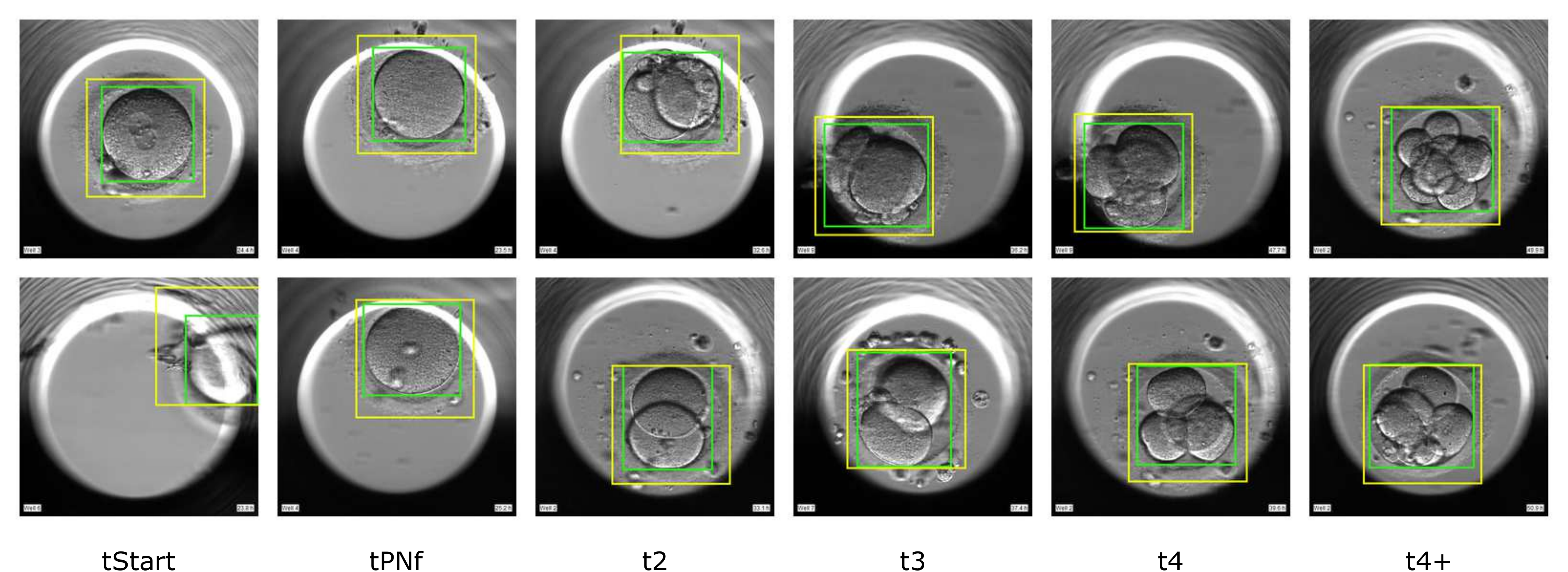}
    \caption{Visualization of detection results by model trained using RL with entropy loss. The green boxes are ground truth bounding boxes and the yellow boxes are network predictions. }
    \label{fig:detection}
\end{figure}

We visualize the detection results of 
a random sample of images
in Figure~\ref{fig:detection}. 
We see that the predicted boxes contain
the region of the ground truth box in almost all images 
and are only fractionally larger than the ground truth boxes.

\begin{table}[htbp]
    \centering
    \caption{Quantitative results of various detector training methods.}
    \begin{tabular}{lccc}
    \toprule
         \textbf{Training Method} & \textbf{Jacc.~Index} & \textbf {Distance} & \textbf{Accuracy}\\
    \midrule
         RL w/ Entropy Loss & \textbf{0.6957} & \textbf{11.58} & \textbf{82.50}\% \\
         RL w/o Entropy Loss & 0.6876 & 18.83 & 75.83\% \\
         Differentiable Sampling & 0.0779 & 228.5 & 71.67\% \\
    \bottomrule
    \end{tabular}
    \label{tab:detection}
\end{table}

\subsection{Embryo Staging}

The baseline model is ResNet-50 applied to raw image resized to $224 \times 224$. 
Our 
method 
(\emph{DetCls}, `detect then classify')
first uses the detector learned in the previous section 
to identify the region of the embryo on $224 \times 224$ resized input.
We experiment with two image cropping methods. 
The first method crops the $112 \times 112$ region on the resized input, 
while the other crops the raw $500 \times 500$ image and resizes it to $224 \times 224$. 
The cropped image with size $112 \times 112$ or $224 \times 224$ 
will then be fed into the same ResNet-50 as the baseline. 
We also try to add LSTM to our DetCls method with $224 \times 224$ crops.

After successfully training the detector in an end-to-end manner, 
we subsequently use the same detector to compare all downstream models, 
The detector is set to test mode to predict only one region. 
We initialize each classifier as a ResNet-50 with pretrained ImageNet weights,
updating all weights using the Adam optimizer 
with a learning rate of $0.001$ and default parameters. 
We apply random rotation augmentation on data. 
The validation data set is used for early stopping 
and all metrics are evaluated on test data.

We report the per-frame accuracy of our raw predictions, 
as well as the per-frame accuracy of the DP predictions (for both objectives). 
We also report the mean absolute error (MAE) and root mean squared error (RMSE) (measured in frames) of the predicted stage transition times after post-processing. 
To better justify these results, we include the result of a naive baseline
that simply 
labels
each frame using the mode stage among all frames captured at the same time in the training set, and predicts the transition time for each stage using the median transition time among all embryo videos in the training set.
Table~\ref{tab:prediction-results} summarizes the results of four models. 

\paragraph{Effect of detection.} 
Two of our single frame DetCls models significantly outperformed the baseline 
before and after post-processing in all metrics. 
Of note, the gain in accuracy due to detection \emph{after post-processing} 
is typically as great or greater than the gains seen in raw accuracy (without post-processing). 
The performance of DetCls112 and DetCls224 is comparable. 
The model with high resolution cropping performs only slightly better after post-processing,
suggesting that the performance gains with respect to the baseline 
are mainly due to removing irrelevant background in the raw input 
and not due to enabling higher-resolution inputs.

\paragraph{Using temporal information.} 
DP post-processing yields an accuracy improvement of $2.69\%$ to $3.61\%$ to all 
three
single-frame models and allows us to generate a monotonic prediction sequence 
to predict the stage transition time. 
DP using Earthmover's distance achieves slightly better performances
on three models (Baseline, DetCls224, DetCls224+LSTM) than DP using likelihood. 
Adding LSTM to DetCls224 further improves the raw accuracy and two metrics (accuracy, MAE) after post-processing. 
The improvement is less significant after post-processing.
This suggests that the DP decoders already encode 
most of the temporal relationships between frames.
\begin{table}
\begin{centering}
\small
\setlength{\tabcolsep}{2pt}
\begin{center}
\caption{
Quantitative results for various architectures and output decoding schemes.
\label{tab:prediction-results}
}
\begin{tabular}{lc|ccc|ccc}
\toprule
& & \multicolumn{3}{c|}{\parbox{0.3\textwidth}{\centering \textbf{DP: label likelihood\\ s.t.~monotonicity}}} & \multicolumn{3}{c}{\parbox{0.27\textwidth}{\centering \textbf{DP: earthmover's distance s.t.~monotonicity}}}\\
\textbf{\textbf{Model}} & \textbf{Raw. Acc.} & \textbf{Accuracy} & \textbf{ MAE} & \textbf{RMSE} & \textbf{ Accuracy} & \textbf{ MAE} &\textbf{RMSE}\\
\midrule
Na\"ive Baseline &  66.87\% & - & - & - & - & 16.42 & 24.34 \\
\midrule
ResNet Baseline  & 83.65\% & 85.54\% & 6.582 & 17.26 & 86.34\% & 6.288 & 16.72  \\
DetCls112 &86.34\% & 89.63\% & 4.594 & 13.38 & 89.13\% & 4.842 & 13.98 \\
DetCls224 &86.18\% & 89.34\% & 4.842 & 14.10 & 89.79\%  & 4.452 & \textbf{12.97} \\
DetCls224+LSTM  & \textbf{88.28\%} & 89.92\% & 4.688 & 14.47 & \textbf{90.23}\% & \textbf{4.370} & 13.67 \\
\bottomrule
\end{tabular} 
\end{center}
\normalsize
\setlength{\tabcolsep}{6pt}
\end{centering}
\end{table}

\begin{figure*}[htbp]
	\centering   
\includegraphics[width=\textwidth]{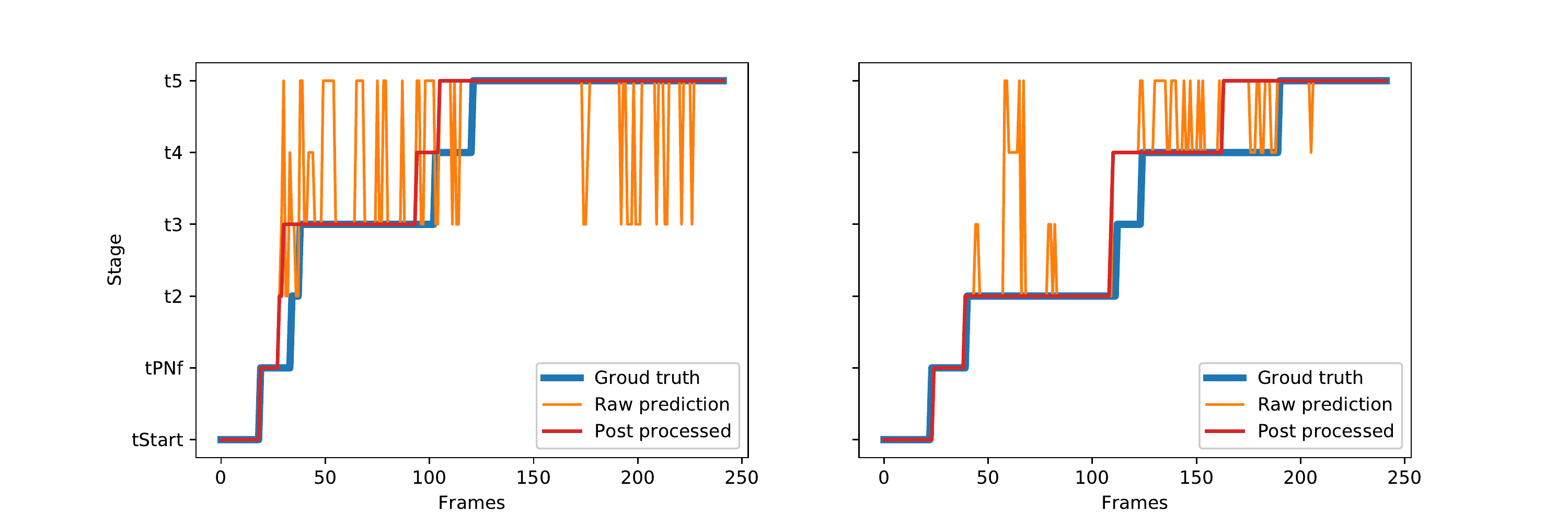}
 \caption{The results of DetCls224 model on two embryo videos, before and after DP post-processing using the earth mover's distance. Post-processing removes the fluctuations in raw predictions to obtain a smooth (monotonic) result.
}
\label{fig:smooth}
\end{figure*}


\section{Related Work}
\label{sec:related}
\subsection{Computer Vision Methodology Papers}

Over the past several years, a variety of papers have made rapid progress
on both single and multiple object detection using convolutinoal neural networks.
Two of the most popular approaches are Faster R-CNN \citep{ren2015faster} and YOLO \citep{redmon2016you}, from which we draw loose inspiration in designing 
our region proposal network for predicting the bounding box.
Traditional object recognition methods are trained on large datasets 
where the true bounding boxes are annotated,
data which is not freely available in many domains, including ours. 
Several previous works seek to address this problem, 
learning weakly-supervised object detection that relies 
only on the image-level class label \citep{bilen2015weakly,bilen2016weakly}. 
Unlike our method, these approaches are not end-to-end trainable with a classifier.

\cite{ba2014multiple}, \cite{sermanet2014attention} and \cite{yeung2016end} use reinforcement learning to learn an attention mechanism for selecting most relevant image regions or video frames for downstream visual recognition tasks.
\citet{jaderberg2015spatial} propose an alternative
method that learns
a geometric transform on the input image, which is fed into a classifier using differentiable bi-linear sampling.
Our work shares a similar idea of localizing the object before classifying it. 

The idea of extending models to include temporal information 
has been explored extensively in recent years.
\cite{Simonyan2014two} used a two-stream architecture applied to a single frame as well as multi-frame optical flow in order to combine spatial and temporal information.
\cite{tripathi2016context} studied techniques for using RNNs to improve frame-level object detection by incorporating context from adjacent frames. They also introduce several additional losses, e.g.~to encourage smoothness in the predictions across adjacent frames. 

\subsection{Embryology Applications Papers}

The problem of predicting embryo annotations from time lapse videos has been addressed in the literature by \cite{Khan2016deep}. 
They use an 8-layer convolutional network 
to count the number of 
cells in an embryo image (up to 5 cells), a related but different setting from ours.
To incorporate temporal information,
they use conditional random fields 
and similarly use dynamic programming 
to enforce monotonicity constraints.
\cite{Milewski} collected annotations of time-lapse morphokinetic data and used principal component analysis and logistic regression analysis to predict pregnancy with an AUC of 0.70.
\cite{Pribenszky} performed a meta-analysis of RCTs comparing use of a morphokinetic algorithm versus single time-point embyro evaluations and found an improved ongoing clinical pregnancy rate with use of the technology.
\cite{zaninovic2018assessing} demonstrated that there is significant variability in some morphokinetic intervals between IVF clinics, suggesting that the parameters used to select embryos may require tuning for each particular clinic.
The human-selected morphorkinetic annotations are in near perfect agreement across repeated exams \citep{embryointerobserver}

Multiple applications of CNNs to embryo assessment were presented 
at the 2018 American Society of Reproductive Medicine Annual Meeting.
\citet{zaninovic2018assessing} analyzed 18,000 images of blastocyts 
using CNNs trained on raw time-lapse images 
and was able to classify the quality of the embryos 
into three morphologic quality grades with 75\% accuracy.
\citet{iwata2018deep} performed a similar analysis to predict good-quality embryos with 80\% accuracy.
\citet{malmsten2018automatic} built a CNN 
using raw time-lapse from images of 11,898 human embryos 
to classify up to the 8-cell stage with 82\% reported accuracy. 
They also reported that the cell-division transition times 
predicted within $5$ frames of when the embryologist annotates the transition
for 91\% of transitions.

To our knowledge, our work is the first to use deep learning 
to predict embryo morphokinetics 
(the above works were published after our work was first made public),
the first to improve performance by localizing the embryo through a weakly-supervised reinforcement learning method, and the first to demonstrate the benefit of incorporating contextual frames via LSTMs.

Beyond embryology, CNN-based classification techniques
have emerged as popular tools in the clinical literature,
with successes such as image-based classification 
of skin lesions \citep{esteva2017dermatologist}
including keratinocyte carcinomas versus benign seborrheic keratoses and malignant melanomas versus benign nevi, 
and the detection of diabetic retinopathy from 
retinal fundus imaging \citep{gulshan2016development}.



\section{Conclusion}
\label{sec:conclusion}
This paper introduced a suite of techniques
for recognizing stages of embryo development,
achieving $90.23\%$ frame-level accuracy.
We also achieve a mean average error for predicting 
stage transition times of $4.370$.
We believe that several directions realizable in future work
could bring this technology to the level of clinical utility.
To begin, our results are achieved using only $510$ embryos for training.
Given that deep learning methods are notable, 
in that performance tends not to saturate quickly with dataset size,
we plan to access a considerably larger dataset for future studies,
to test the limits of our current methodology.
Additionally, our models are initialized by using pretrained weights 
from an ImageNet classifier originally on full color photographs.
We suspect that transfer from a comparably large dataset of more relevant images
(gray-scale microscopy) might yield additional gains. 
Identifying such a dataset for transfer remains a challenge.
Moreover, even if we can access such a dataset of unlabeled images,
deciding upon a (possibly unsupervised) objective for the source task 
could poses an interesting research problem.
Additionally, we plan to extend our experiments 
to predict not only the $6$ stages useful for current embryo selection heuristics 
but to predict all $15$ stages of development. 
And finally, we hope to use the models learned 
from the morphokinetic prediction as themselves a source task, 
fine-tuning the models to the more pressing downstream problem
of assessing implantation potential directly. 
We note that assessing the viability of an embryo 
represents an interesting off-policy learning problem. 
Outcomes are only observed for those embryos that implanted. 
Success on this task may require not only representation learning,
but also estimating counterfactual quantities.

\bibliographystyle{plainnat}
\bibliography{refs}

\end{document}